\documentclass{easychair}

\usepackage{doc}
\usepackage{makeidx}
\usepackage[leqno,fleqn,intlimits]{amsmath}
\usepackage{graphicx}
\usepackage{color}
\usepackage{float}
\usepackage{url}
\usepackage[utf8]{inputenc}
\usepackage[T1]{fontenc}

\definecolor{red_ex}{rgb}{0.7,0,0}
\definecolor{blue_ex}{rgb}{0,0,0.7}

\begin{document}


\title{ Sentence Compression in Spanish driven by Discourse Segmentation and Language Models}


\titlerunning{ Sentence Compression by Discourse Segmentation and LM}

%
\author{
    Alejandro {\sc Molina}\inst{1}
\and
Juan-Manuel {\sc Torres-Moreno}\inst{1}$^,$\inst{2}
\and
   Iria {\sc da Cunha}\inst{3}
\and
    Eric {\sc SanJuan}\inst{1}
\and
    Gerardo {\sc Sierra}\inst{4}\\
}

\institute{
  Laboratoire Informatique d'Avignon,\\
  BP 91228 84911, 
  Avignon, Cedex 09, France\\
  \email{\{alejandro.molina,juan-manuel.torres,eric.sanjuan\}@univ-avignon.fr}
\and
   \'Ecole Polytechnique de Montr\'eal,\\ CP. 6128 succursale Centre-ville,
   Montréal, Québec, Canada\\
\and
   Universitat Pompeu Fabra,
   Barcelona, Spain.\\
   \email{iria.dacunha@upf.edu}
\and
   Instituto de Ingenier\'a, UNAM
   Mexico, DF.\\
   \email{gerardo.sierra@iingen.unam.mx}
 }

\authorrunning{\sc Molina et al.}

\clearpage

\maketitle

\begin{abstract}
Previous works demonstrated that Automatic Text Summarization (ATS) by sentences extraction may be improved using sentence compression. 
In this work we present a sentence compressions approach guided by level-sentence discourse segmentation and probabilistic language models (LM).
The results presented here show that the proposed solution is able to generate coherent summaries with grammatical compressed sentences.
The approach is simple enough to be transposed into other languages.
\end{abstract}

\section{Introduction}

Automatic Text Summarization (ATS) is indispensable to cope with ever increasing volumes of valuable information. 
An abstract is by far the most concrete and most recognized kind of text condensation \cite{ANS:79}.
Sentences extraction allows to generate summaries by extraction sentences \cite{luhn:58,edmundson:69,torres:11}. 

Sentence compression can be used to improve extract summarization \cite{knight:00,molina:linguamatica:10}. 
Previous works suggest that sentence segmentation could be helpful in sentence compression generation \cite{sporleder:05}.

In this work we present an new automatic sentence compression generation approach.
First sentences are segmented using a discourse segmenter and then, compression candidates are generated.
Finally, the best candidate i.e., the most grammatical one, is selected based on its probability as a sequence in a Language Model.

We organized the rest of the paper as follows. 
Firstly, in section \S\ref{sec:state} we recall the mains concepts of sentence compression. 
Then, we present in \S\ref{sec:SC} our compression candidates generation approach. 
Compression candidates evaluation is introduced in \S\ref{sec:Score}.
Experimental results are showed in \S\ref{sec:resultats}.
Finally, section  \S\ref{sec:conclusion} presents conclusions and future work.

\section{Sentence compression}
\label{sec:state}
Sentence compression can be considered as a summarization at the sentence level.
Sentence compression task is defined as follows:

``Consider an input sentence as a sequence of $n$ words $W = (w_1, w_2, ..., w_n)$.
An algorithm may drop any subset of these words.
The words that remain (order unchanged) form a compression'' \cite{knight:02}.

There are interesting algorithms to determine the removal of words in a sentence but humans tend also to delete long phrases in an abstract \cite{pitler:10}.

Recent studies have found good results by concentrating on clauses, instead of isolated words.
In \cite{steinberger:06} an algorithm first divides sentences into clauses prior to any elimination and then, compression candidates are scored based on Latent Semantic Analysis proposed in \cite{deerwester:90}.
However, no component to mitigate grammaticality issues is included in this algorithm \cite{steinberger:06}. 
Although the results of this last work are in general good, in some cases the main subject of the sentence is removed.
The authors attempted to solve this issue by including features in a machine learning approach \cite{steinberger:07}.

As an alternative to clauses, some studies explore discourse structures to tackle the sentence compression task.
Discourse chunking \cite{sporleder:05} is an alternative to discourse parsing, thereby, showing a direct application to sentence compression.
The authors of this last work plausibly argued that, while discourse parsing at document-level stills poses a significant challenge, sentence-level discourse chunking could represent an alternative in languages with limited full discourse parsing tools.
In addition, some sentence-level discourse models have shown accuracies comparable to human performance \cite{soricut:03}.




\section{Compression Candidates Generation}
\label{sec:SC}


In this work, we use a sentence-level discourse segmentation approach.
Formally, ``Discourse segmentation is the process of decomposing discourse into Elementary Discourse Units (EDUs), which may be simple sentences or clauses in a complex sentence, and from which discourse trees are constructed'' \cite{tofiloski:09}.
Discourse segmentation is only the first stage for discourse parsing (the others are detection of rhetorical relations and building of discourse trees).
However, we can consider segmentation at the sentence level in order to identify segments to be eliminated in the sentence compression task.
This decomposition of a sentence into EDUs using only local information is called shallow discourse segmentation.
In \cite{molina:micai:11}, the authors use a discourse segmenter in order to segment sentences in spanish.
The discourse segmenter is described in \cite{dacunha:10} and is based in the Rhetorical Structure framework \cite{mann:87}.

We propose that compression candidates be generated by deleting some discourse segments from the original sentence.
Let be a sentence $S$ the sequence of its $k$ discourse segments: $S=(s_1, s_2, . . . , s_k)$.
A candidate, $CC_{i}$, is a subsequence of $S$ that preserves the original order of the segments.
The original sentence always form a candidate, i.e., $CC_0=S$, 
this is convenient because sometimes there is no shorter grammatical version of the sentence,
especially in short sentences that conform one single EDU.
Since we do not consider the empty subsequence as a candidate, there are $2^k-1$ candidates.
Furthermore, since we rarely have more than 5 discourse segments in a sentence, usually we create between 1 and 31 candidates,
this, dramatically reduces the solution space given that $k << n$.
The compression candidates are constructed using a $O(2^k)$ binary counter.
In Example 1 we show all the candidates associated to a sentence extracted from our corpus.

\begin{quote}

Example 1. \\
$CC_0$:[Adem\'as ella particip\'o ese mismo a\~no en el concierto en tributo a Freddie Mercury,][hablando acerca de la prevenci\'on necesaria][para combatir el SIDA.]\footnote{English translation: [Also she participated that year in the concert in tribute to Freddie Mercury, ][talking about prevention needed][to fight AIDS.]}
\\
$CC_1$:[hablando acerca de la prevenci\'on necesaria][para combatir el SIDA.]
\\
$CC_2$:[Adem\'as ella particip\'o ese mismo a\~no en el concierto en tributo a Freddie Mercury,][para combatir el SIDA.]
\\
$CC_3$:[para combatir el SIDA.]
\\
$CC_4$:[Adem\'as ella particip\'o ese mismo a\~no en el concierto en tributo a Freddie Mercury,][hablando acerca de la prevenci\'on necesaria]
\\
$CC_5$:[hablando acerca de la prevenci\'on necesaria]
\\
$CC_6$:[Adem\'as ella particip\'o ese mismo a\~no en el concierto en tributo a Freddie Mercury,]

\end{quote}

	
\section{Compression Candidates Scoring with Language Model}
\label{sec:Score}
A Language Model (LM) estimates the probability distribution of natural language.
Statistical language modeling \cite{chen:99,manning:99} is a technique widely used to assign a probability to a sequence of words.
We assume that good compression candidates must have a high probability as sequences in a LM.
In general, for a sentence $W = (w_1, w_2, . . . , w_n)$,  the probability of $W$ is:
\begin{eqnarray}
	P(w_1^n) = P(w_1)P(w_2 | w_1)P(w_3 | w_1^2) & \nonumber \\
	...P(w_n | w_1^{n-1}) &
\end{eqnarray}
Where $w_a^{b}=(w_a, . . . , w_b)$.
The probabilities in a LM are estimated counting sequences from a corpus.
Even though we will never be able to get enough data to compute the statistics for all possible sentences,
we can base our estimations using big corpora and interpolation methods.
In our experiments we use a big corpus with 1T words\footnote{LDC Catalog No.: LDC2009T25 ISBN:  1-58563-525-1
} to get the sequences counts and 
a LM interpolation based on Jelinek-Mercer smoothing \cite{chen:99}:

\begin{eqnarray}
\label{eq:interpolation}
 P_{\textrm{interp}}(w_i | w_{i-n+1}^{i-1}) & = & \nonumber \\
  \lambda_{w_{i-n+1}^{i-1}} 
   P_{\textrm{max likelihood}}(w_i | w_{i-n+1}^{i-1})  &+ &\nonumber \\
  (1-\lambda_{w_{i-n+1}^{i-1}}) P_{\textrm{interp}}(w_i | w_{i-n+2}^{i-1})
\end{eqnarray}

In equation (\ref{eq:interpolation}) the maximum likelihood estimate of a sequence is interpolated with the smoothed 
lower-order distribution.

For a given candidate, 
\begin{equation}
CC_i = (s_{1}, ... , s_{k}) = (w_{1}, ... ,w_{n}),
\end{equation}
we assign a LM based score ($Score_{g}$) based on its probability as in equation (\ref{eq:score}).
For our experiments we use the Language Modeling Toolkit SRILM\footnote{Avaliable at 
\url{http://www-speech.sri.com/projects/srilm/download.html}
} \cite{stolcke:02}.

\begin{equation}
\label{eq:score}
Score_{g}(CC_i)  = \frac{P_{\textrm{interp}}(CC_i) }{ n}
\end{equation}






\section{Experimental Results}
\label{sec:resultats}

For the experiments,
two annotators were required to compress each sentence following the instructions in \cite{molina:micai:11}.
The corpus contains four sub-corpora: 
 Wikipedia sections, brief news, scientific abstracts and short stories.
Each sub-corpus has 20 texts composed of no more than 50 phrases each one (1939 tokens).
We have randomly selected eight documents for evaluation, two of each sub-corpus.

We have generated abstract summaries selecting the best compression candidate of each sentence
considering two different approaches:
\begin{enumerate}
\item All system: Selecting the best scored candidate for each sentence.
\item First system: Selecting the best candidate from those that include the first segment.
\end{enumerate}

For comparison we have created Random system: a baseline system which applies a random compression.
Random system eliminates some words of a given sentence at the same rate of human annotators.

After compressions, three judges (different from annotators) read the eight summaries.
The judges do not know the source of the final summary.
They mark each sentence in the final summary if they found it grammatically incorrect in the context.
In addition, they evaluate the global coherence of summaries after compressions.
Coherence of summaries is scored with a categorical variable:
a value of -1 is assigned for incoherent summaries, 0 if some coherences are found and +1 for coherent productions.
The Compression Rate (CR) is defined as the proportion of content eliminated from the original document.
It says how much of the content was eliminated.

In Table~\ref{tab:gramm}, we compare our systems and the two human annotators.
The results confirm that scoring compressions before producing a summary improves the text quality in summaries with compressed sentences.
It is very surprising that, for Human$_{2}$, the number of compressions judged grammatically incorrect is greater than that of our systems.
May be Human$_{2}$ misunderstood the compression instructions.
However, considering as limits the results of Human$_{1}$ and Random baseline system we consider that the proportion of bad formed sentences is very low.
We confirmed our initial intuition that preserving the first segment tends to save the main subject in most of the cases.
The introduction of this simple heuristic in the First system improves the grammar quality of productions.

Table~\ref{tab:rouge} shows the result of comparing our systems using the two human summaries as references.
We wanted to evaluate the content quality of summaries with the ROUGE package \cite{lin:2003}.
ROUGE is used to evaluate summaries because some results show that it correlates well with human judgements \cite{lin:2004rpa}. 
Results in Table~\ref{tab:rouge} are opposite to what was expected.
We assumed that Random system would have worst results with respect to First system.
Looking at the judgments of coherence and grammar, made by humans, in Table~\ref{tab:gramm}, we expected the same positions of the systems. 
However, in Table~\ref{tab:rouge} we see that the best value of ROUGE, using human-made summaries with compressed sentences as references, is for the Random system.
Other than that, again, First system overcomes All system.

As an alternative, we compare the divergence of texts with respect to the original uncompressed text using the
FRESA package\footnote{
\url{http://lia.univ-avignon.fr/fileadmin/axes/TALNE}
}
 \cite{torres:10c,saggion:10,torres:10poli}. 
The FRESA score F$_M$ asses the summaries qualities.
Lower values of F$_M$ means significant difference whit respect to the original text (i.e. more radical compressions).
The results of divergence tests of summaries is showed in Table~\ref{tab:fresa}. 
Results in Table~\ref{tab:fresa} are interesting.
Considering F$_M$ values, we see that First system is found more related to Human$_{1}$ than Human$_{2}$ by FRESA and
Human$_{2}$ is closer to Random system performance.
These values are congruent with the coherence and grammar judgments showed in Table~\ref{tab:gramm}.
The F$_M$ value of All system suggests that it is the most aggressive approach.

For all tables we use the following notation about the sources: 
All system=all compression candidates, 
First system=candidates including the first segment, 
Random system=random compression (baseline), 
Human$_{1,2}$=human compressions.

\begin{table}[h]
\begin{center}
  \begin{tabular}{|lccc|}
  \hline
  					 & {   Avgerage}			& {   } 	&   Agrammatical \\
{   Algorithm}	 & {   CR} 			& {  Coherence } 			&    compressions\\
					& {   (\%)}					 & {   } 		&    (\%) \\
  \hline
 	All  		&   30.50 &    +0.37  &   8.12\\
 	First 	&    18.80 &    +0.62  &   6.98\\
 	Random 	&    22.97 &   -0.50  &   76.60\\
 	Human$_1$ 	&    22.34 &   +1.00  &   0.00\\
 	Human$_2$ 	&    15.98 &   +0.75  &   20.68\\
 \hline
 \end{tabular}
 \caption{Gramaticallity tests.}
 \label{tab:gramm}
 \end{center}
\end{table}

\begin{table}[h]
\begin{center}
  \begin{tabular}{|lccc|}
  \hline
{   Algorithm}	  & {   ROUGE-1 } 		& {   ROUGE-2 } 		&   ROUGE-SU4  \\
  \hline
 	All  &   0.6999 &   0.6041 &   0.5860\\
 	First  &    0.8069 &    0.6897 &    0.6775\\
 	Random &   0.8880 &   0.7085 &    0.7257\\
 \hline
 \end{tabular}
 \caption{Content tests using ROUGE package. The summaries references were the two humans compressions. }
 \label{tab:rouge}
 \end{center}
\end{table}

\begin{table}[h]
\begin{center}
 \tabcolsep = 0.8\tabcolsep
  \begin{tabular}{|lcccc|}
  \hline
{  Algorithm}	  & {  F$_1$ } & {  F$_2$ } & {  F$_4$ }&{  F$_M$ }  \\
  \hline

 	All  		&   0.9197 	&   0.9124 	&   0.9078	&    0.9133\\
 	First  		&   0.9461 	&   0.9472 	&   0.94512	&   0.9461\\
 	Random	&   0.9593 	&   0.9536 	&   0.9535 	&   0.9555\\
 	Human$_1$ 	&   0.9460 	&   0.9427 	&   0.9372 	&   0.9420\\
 	Human$_2$ 	&   0.9594 	&   0.9562 	&   0.9547 	&   0.9567\\
 \hline
 \end{tabular}
 \caption{Divergence tests by using FRESA without references. F$_M$ is an statistical mean of 1-grammes (F$_1$), 
2-grammes (F$_2$) and SU4-grammes (F$_4$).}
 \label{tab:fresa}
 \end{center}
\end{table}

\section{Conclusions and future work}
\label{sec:conclusion}

In this work we have introduced the concept of Sentence Compression driven by Discourse Segmentation and Language Models.
We have found that using Probabilistic Language Models can be helpful for evaluation of compressions candidates.
The results in Spanish presented in this paper are very encouraging.
We believe that this approach is independent enough of the language to be transposed into other languages such as English or French.
In future work we aim to improve the score 
(\ref{eq:score}) adding content restrictions.

Evaluation of compressed sentences and summaries with compressions is still a challenge in languages other than English that do not have reference corpora.
We think that more studies are necessary in order to evaluate if ROUGE or FRESA are good methods for compressed text evaluations.


\section*{Acknowledgments}

This work was partially supported by {\sl Consejo Nacional de Ciencia y Tecnolog\'ia} (CONACYT) México, grant number 211963.

\bibliographystyle{plain}
\bibliography{biblio}

\begin{thebibliography}{10}

\bibitem{ANS:79}
ANSI.
\newblock {\em {American National Standard for Writing Abstracts Z39-14}}.
\newblock ANSI, New York, 1979.

\bibitem{chen:99}
S.F. Chen and J.~Goodman.
\newblock An empirical study of smoothing techniques for language modeling.
\newblock {\em Computer Speech \& Language}, 13(4):359--393, 1999.

\bibitem{dacunha:10}
Iria da~Cunha, Eric SanJuan, Juan-Manuel Torres-Moreno, Marina Lloberes, and
  Irene Castellón.
\newblock Discourse segmentation for spanish based on shallow parsing.
\newblock In Grigori Sidorov, Arturo Hernández~Aguirre, and Carlos
  Reyes~García, editors, {\em Advances in Artificial Intelligence}, volume
  6437 of {\em Lecture Notes in Computer Science}, pages 13--23. Springer
  Berlin / Heidelberg, 2010.
\newblock 10.1007/978-3-642-16761-4-2.

\bibitem{deerwester:90}
S.~Deerwester, S.~T. Dumais, G.~W. Furnas, T.~K. Landauer, and R.~Harshman.
\newblock {Indexing by Latent Semantic Analysis}.
\newblock {\em Journal of the American Society for Information Science},
  41(6):391--407, 1990.

\bibitem{edmundson:69}
H.~P. Edmundson.
\newblock {New Methods in Automatic Extraction}.
\newblock {\em Journal of the Association for Computing Machinery},
  16(2):264--285, 1969.

\bibitem{knight:00}
Kevin Knight and Daniel Marcu.
\newblock Statistics-based summarization -- step one: Sentence compression.
\newblock In {\em Proceedings of the 17th National Conference on Artificial
  Intelligence and 12th Conference on Innovative Applications of Artificial
  Intelligence}, pages 703--710, Austin, TX, Etats-Unis, 2000.

\bibitem{knight:02}
Kevin Knight and Daniel Marcu.
\newblock {Summarization beyond sentence extraction: a probabilistic approach
  to sentence compression}.
\newblock {\em Artificial Intelligence}, 139(1):91--107, Juillet 2002.

\bibitem{lin:2004rpa}
Chin-Yew Lin.
\newblock {ROUGE: A Package for Automatic Evaluation of Summaries}.
\newblock In Marie-Francine Moens and Stan Szpakowicz, editors, {\em
  Proceedings of the Workshop Text Summarization Branches Out (ACL'04)}, pages
  74--81, Barcelone, Espagne, Juillet 2004. ACL.

\bibitem{lin:2003}
Chin-Yew Lin and Eduard Hovy.
\newblock {Automatic Evaluation of Summaries using N-gram Co-occurrence
  Statistics}.
\newblock In {\em Proceedings of the 2003 Conference of the North American
  Chapter of the Association for Computational Linguistics on Human Language
  Technology (NAACL'03)}, volume~1, pages 71--78, Edmonton, Canada, 2003. ACL.

\bibitem{luhn:58}
H.P. Luhn.
\newblock {The Automatic Creation of Literature Abstracts}.
\newblock {\em IBM Journal of Research and Development}, 2(2):159--165, 1958.

\bibitem{mann:87}
W.~C. Mann and S.~A. Thompson.
\newblock {\em {Rhetorical Structure Theory: A Theory of Text Organization}}.
\newblock Information Sciences Institute, Marina del Rey, 1987.

\bibitem{manning:99}
Christopher~D. Manning and Hinrich Sch{\"u}tze.
\newblock {\em {Foundations of Statistical Natural Language Processing}}.
\newblock MIT Press, Cambridge, 1999.

\bibitem{molina:linguamatica:10}
Alejandro Molina, Iria da~Cunha, Juan-Manuel Torres-Moreno, and Patricia
  Velazquez-Morales.
\newblock La compresi\'on de frases: un recurso para la optimizaci\'on de
  resumen autom\'atico de documentos.
\newblock {\em Linguam\'atica}, 2(3):13--27, 2010.

\bibitem{molina:micai:11}
Alejandro Molina, Torres-Moreno Juan-Manuel, Eric~SanJuan Eric, Iria da~Cunha,
  Gerardo Sierra, and Patricia Vel\'azquez-Morales.
\newblock {Discourse Segmentation for Sentence Compression}.
\newblock In {\em {Proceedings of the Mexican International Conference on
  Artificial Intelligence (MICAI'11)}}, page 10 pages, Puebla, Mexico, 2011.
  Springer-Verlag.

\bibitem{pitler:10}
Emily Pitler, Annie Louis, and Ani Nenkova.
\newblock Automatic evaluation of linguistic quality in multi-document
  summarization.
\newblock In {\em Proceedings of the 48th Annual Meeting of the Association for
  Computational Linguistics}, ACL '10, pages 544--554, Stroudsburg, PA, USA,
  2010. Association for Computational Linguistics.

\bibitem{saggion:10}
Horacio Saggion, Juan-Manuel Torres-Moreno, Iria da~Cunha, and Eric SanJuan.
\newblock {Multilingual summarization evaluation without human models}.
\newblock In {\em Proceedings of the 23rd International Conference on
  Computational Linguistics: Posters (COLING'10)}, pages 1059--1067, Beijing,
  Chine, 2010. ACL.

\bibitem{soricut:03}
Radu Soricut and Daniel Marcu.
\newblock Sentence level discourse parsing using syntactic and lexical
  information.
\newblock In {\em HLT-NAACL}, 2003.

\bibitem{sporleder:05}
Caroline Sporleder and Mireille Lapata.
\newblock {Discourse chunking and its application to sentence compression}.
\newblock In {\em Proceedings of the conference on Human Language Technology
  and Empirical Methods in Natural Language Processing}, pages 257--264.
  Association for Computational Linguistics, 2005.

\bibitem{steinberger:06}
Josef Steinberger and Karel Jezek.
\newblock {\em Sentence Compression for the LSA-based Summarizer}, pages
  141---148.
\newblock 2006.

\bibitem{steinberger:07}
Josef Steinberger and Roman Tesar.
\newblock Knowledge-poor multilingual sentence compression.
\newblock In {\em 7th Conference on Language Engineering (SOLE'07)}, pages
  369--379, Cairo, Egypt, 2007.

\bibitem{stolcke:02}
A.~Stolcke.
\newblock Srilm -- an extensible language modeling toolkit.
\newblock In {\em Intl. Conf. on Spoken Language Processing}, volume~2, pages
  901--904, Denver, 2002.

\bibitem{tofiloski:09}
Milan Tofiloski, Julian Brooke, and Maite Taboada.
\newblock A syntactic and lexical-based discourse segmenter.
\newblock In {\em ACL-IJCNLP}, 2009.

\bibitem{torres:10c}
Juan-Manuel Torres-Moreno.
\newblock Fresa: a framework for evaluating summaries automatically.
\newblock rapport technique, Université d'Avignon et des Pays de Vaucluse,
  Avignon, France, 2010.

\bibitem{torres:11}
Juan-Manuel Torres-Moreno.
\newblock {\em {Résumé automatique de documents: une approche statistique}}.
\newblock {Hermès-Lavoisier, Paris}, 2011.

\bibitem{torres:10poli}
Juan-Manuel Torres-Moreno, Horacio Saggion, Iria da~Cunha, and Eric SanJuan.
\newblock {Summary Evaluation With and Without References}.
\newblock {\em {Polibits: Research journal on Computer science and computer
  engineering with applications}}, 42:13--19, 2010.

\end{thebibliography}

\end{document}